\pdfoutput=1
\documentclass[journal]{IEEEtai}

\usepackage[colorlinks,urlcolor=blue,linkcolor=blue,citecolor=blue]{hyperref}

\usepackage{amsmath,amssymb,amsthm}
\theoremstyle{plain}

\newtheorem*{theorem*}{Theorem}

\theoremstyle{remark}

\theoremstyle{definition}

\newtheorem{mydef}{\bf{Definition}}

\usepackage[dvipsnames]{xcolor}
\usepackage{amsmath}
\usepackage{mathtools}
\usepackage{ulem}
\usepackage[noadjust]{cite}
\usepackage{graphicx}
\usepackage{multicol}
\usepackage{array}
\usepackage{multirow}
\usepackage{latexsym}
\usepackage{times}
\usepackage{booktabs}
\usepackage{float}
\usepackage{placeins}
\usepackage{mathtools, amsfonts, amssymb, amsthm, bm}
\usepackage[english]{babel}
\usepackage[figurename=Fig.,font={small,it}]{caption}
\usepackage{subcaption}
\usepackage{siunitx}
\usepackage[utf8]{inputenc}
\usepackage[T1]{fontenc}
\usepackage{authblk}
\usepackage[numbers]{natbib}
\usepackage{algorithm}
\usepackage{algorithmic}
\usepackage{tabularx}
\usepackage{tikz}
\usetikzlibrary{arrows.meta, decorations.markings, decorations.pathreplacing}
\usepackage{svg}

\newcommand{\R}{\mathbb{R}}
\newcommand{\E}{\mathbb{E}}

\newcommand{\loss}{\mathcal{L}}
\newcommand{\dataset}{\mathcal{S}}
\newcommand{\dataD}{\mathcal{D}}

\begin{document}

\title{SAM-on-the-Curve: Sharpness-Aware Mode Connectivity for Robust Weight-Space Interpolation}


\author{Alejandro Calatrava, Xu Zhang and Ren Wang}

\maketitle

\begin{abstract}
Deep neural networks that are independently trained to similar performance can be connected by low-loss parametric curves in weight space, a phenomenon known as Mode Connectivity (MC). This geometric property underpins practical techniques such as weight averaging, model ensembling, and model merging. We argue that low-loss connectivity is an incomplete geometric criterion: it controls loss only along a one-dimensional trajectory while leaving the surrounding weight-space neighborhood unconstrained, so the optimized curve may traverse sharp ridges that become fragile under distribution shift. We therefore reformulate mode connectivity as a neighborhood-robust path optimization problem, seeking a curve whose entire local neighborhood maintains low loss. We propose Sharp Mode Connectivity (SMC), which applies a first-order sharpness-aware approximation to the resulting minimax functional, enforcing flatness along the entire curve rather than only on it. We derive a practical optimization algorithm for connectivity paths under this sharpness-aware objective. Under severe blur corruptions from CIFAR-10-C, SMC achieves up to 6.09\% absolute accuracy improvement over standard MC. Remarkably, SMC produces negative loss barriers, meaning that models obtained at interior points of the optimized path can outperform the average endpoint loss. These results, validated across ResNet-18, VGG16-BN, and ViT-Tiny on CIFAR-10 and ImageNet-100, establish path-wise flatness as a practical principle for robust weight-space interpolation.
\end{abstract}

\begin{IEEEImpStatement}
As deep learning is increasingly deployed in safety-critical applications such as autonomous driving, medical diagnosis, and scientific instrumentation, the ability to guarantee robust generalization under real-world distribution shifts becomes a societal imperative, not merely an academic benchmark. This work contributes a principled geometric framework that enables the construction of inherently robust model ensembles and weight-averaged models by ensuring that the entire weight-space path connecting independently trained solutions lies within flat, low-loss valleys. For practitioners, this means more reliable predictions when operating conditions deviate from training data, directly impacting the trustworthiness of AI-assisted decisions in healthcare, infrastructure monitoring, and environmental sensing. For the research community, this work bridges two influential but previously disconnected theoretical perspectives (Mode Connectivity and Sharpness-Aware Minimization), opening new avenues in path-wise geometric regularization and, potentially, federated model merging (where client models must be combined through weight-space operations), as well as a deeper understanding of how architectural inductive biases shape the geometry of the loss landscape.
\end{IEEEImpStatement}



\section{Introduction}\label{sec:intro}

\IEEEPARstart{W}{eight-space} interpolation has become a central operation in modern deep learning. Stochastic Weight Averaging (SWA)~\cite{izmailov2018averaging} averages SGD iterates along training trajectories to find wider optima. Fast Geometric Ensembling (FGE)~\cite{modeConnectivity} collects diverse checkpoints along low-loss curves discovered by cyclical learning rates. Model soups~\cite{wortsman2022model} combine independently fine-tuned parameter vectors to improve accuracy without increasing inference cost. All of these methods rely on combining or interpolating parameters across trained solutions in weight space, so the geometric structure of the region connecting these solutions can influence the robustness of the resulting models.

Mode Connectivity (MC)~\cite{modeConnectivity, draxler2018essentially} provides the foundational path-finding framework: given two independently trained solutions $w_1, w_2 \in \R^d$, MC optimizes a parametric curve $\phi_\theta$ connecting them by minimizing the expected loss along the path. However, this low-loss connectivity is an incomplete geometric criterion because it controls the loss strictly on the one-dimensional trajectory while leaving the surrounding weight-space neighborhood unconstrained. Consequently, the optimized path may thread through narrow, sharp ridges where the loss rises steeply under perturbation, making it fragile under distribution shift. We therefore reformulate mode connectivity as a neighborhood-robust path optimization problem, seeking a curve whose entire local neighborhood maintains low loss.

\begin{figure}[htbp]
    \centering
    \includegraphics[width=0.95\linewidth]{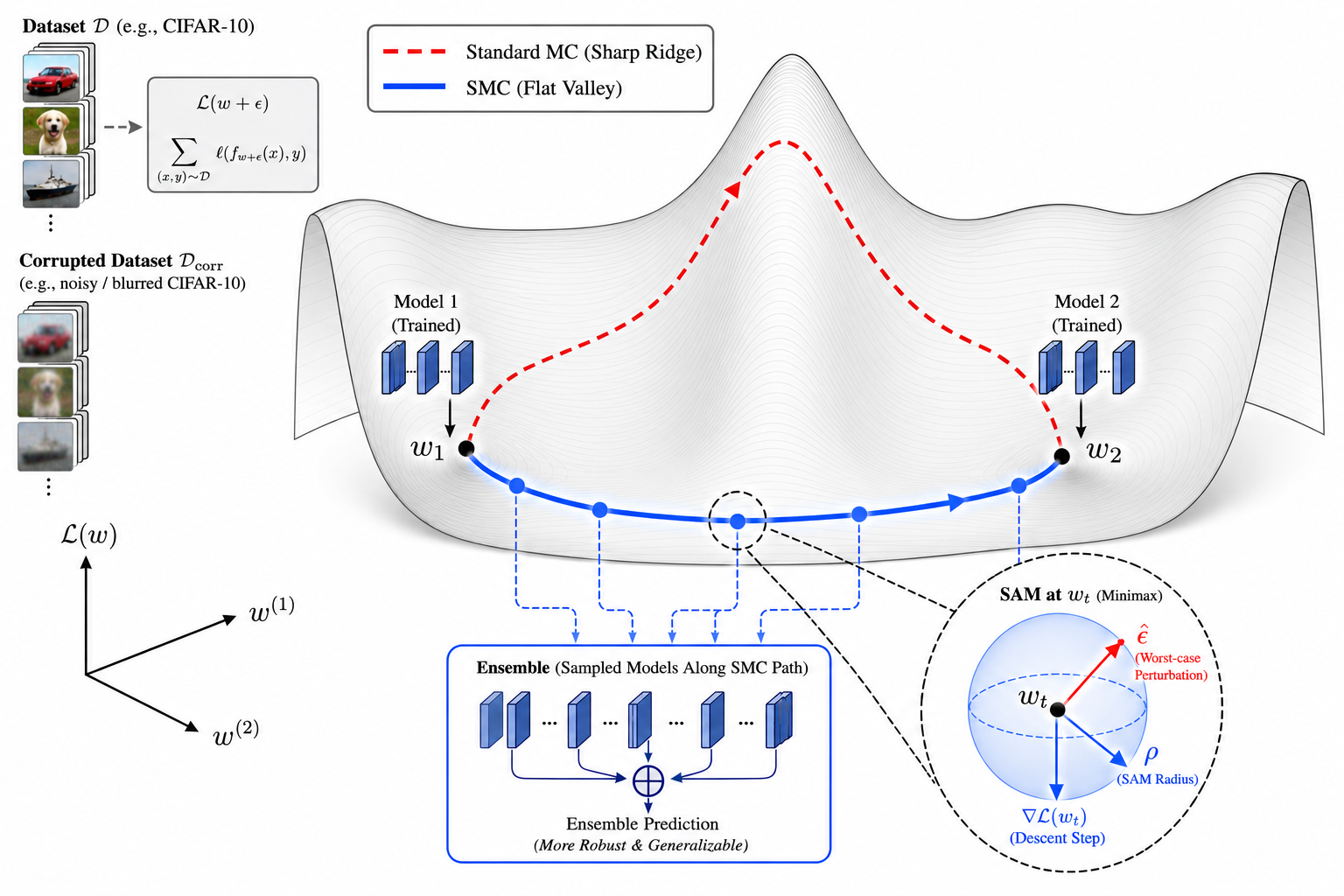}
    \caption{Conceptual overview of SMC. Two independently trained solutions $w_1$ and $w_2$ are connected by a standard MC path (red, dashed), which may traverse sharp ridges, and an SMC path (blue, solid), which is encouraged to pass through flatter regions by the sharpness-aware perturbation objective of radius $\rho$.}
    \label{fig:concept}
\end{figure}

We propose \textit{Sharp Mode Connectivity} (SMC), which replaces the pointwise loss at each curve location with the worst-case loss inside a surrounding perturbation neighborhood, so the whole curve is driven into a flat region of the loss landscape (Fig.~\ref{fig:concept}; formalized in Section~\ref{sec:method}). Empirically, this makes the loss profile along the curve far more uniform: on CIFAR-10 with ResNet-18, the variance of the loss across points sampled along the SMC path is $0.0021$, versus $0.056$ for standard MC (a $26.7\times$ reduction), reflecting a nearly flat, valley-like profile rather than a sharp point spike. Under severe distribution shifts from the CIFAR-10-C benchmark~\cite{hendrycks2019robustness}, the path midpoint achieves up to $+6.09\%$ absolute accuracy improvement over standard MC. We focus on the midpoint here to evaluate the robustness of the connectivity, whereas peak accuracy is reported later when seeking the single best model along the path. Furthermore, SMC produces \emph{negative loss barriers}.

Our principal contributions are:
\begin{itemize}
    \item \textbf{Neighborhood-robust mode connectivity.} We identify a limitation of conventional mode connectivity: minimizing loss on a one-dimensional interpolation trajectory leaves the surrounding weight-space neighborhood unconstrained. We formulate SMC as a minimax path functional that seeks low-loss neighborhoods along the connectivity curve.
    \item \textbf{SAM-on-the-Curve optimization.}  We derive a practical stochastic algorithm that optimizes Bézier control points using first-order sharpness-aware perturbations at sampled curve locations, extending sharpness-aware optimization from isolated parameter vectors to continuous weight-space trajectories.
    \item \textbf{Geometric and robustness analysis.} Across CNN and Transformer architectures, we show that sharpness-aware path optimization reduces both longitudinal path-loss variation and local neighborhood sensitivity. Under CIFAR-10-C distribution shifts, these geometric changes are associated with improved interpolation robustness and path-interior solutions that can outperform the endpoint average.
\end{itemize}

The remainder of this paper is organized as follows. Section~\ref{sec:related} reviews related work. Section~\ref{sec:method} introduces the necessary preliminaries on Mode Connectivity and SAM, identifies the research gap, and presents the proposed SMC framework and the SAM-on-the-Curve algorithm. Section~\ref{sec:experiments} reports comprehensive experimental results. Section~\ref{sec:conclusion} concludes and discusses future directions.

\section{Related Work}\label{sec:related}

\subsection{Generalization and Loss Landscape Geometry}\label{subsec:related_landscape}
Understanding why overparameterized DNNs generalize despite having the capacity to memorize random labels~\cite{zhang2017understanding} has motivated a shift toward geometric explanations of generalization. A central finding is that the \emph{shape} of minima in the loss landscape, rather than merely the loss value, is among the strongest predictors of test performance~\cite{jiang2020fantastic}.

Hochreiter and Schmidhuber~\cite{hochreiter1997flat} first established that flat minima, whose loss remains stable under parameter perturbations, are associated with shorter description lengths and better generalization. Keskar et al.~\cite{keskar2017large} provided complementary empirical evidence showing that large-batch SGD converges to sharp, narrow minima with degraded test accuracy, whereas small-batch training favors flatter regions. Neyshabur et al.~\cite{neyshabur2017exploring} formalized norm-based generalization bounds connecting parameter geometry to the gap between training and test error, further grounding the flatness and generalization link in learning theory.

Li et al.~\cite{li2018visualizing} introduced filter-normalized loss landscape visualization, enabling meaningful geometric comparisons across architectures and training regimes. Their analysis confirmed that architectural choices such as skip connections (e.g., ResNets~\cite{he2016resnet}) produce markedly smoother loss surfaces than architectures without residual paths (e.g., VGG~\cite{simonyan2015vgg}), a finding that influences both optimization dynamics and the feasibility of weight-space interpolation methods discussed below.

\subsection{Sharpness-Aware Optimization}\label{subsec:related_sam}

Motivated by the flatness and generalization connection, Foret et al.~\cite{foret2020sharpness} proposed \textit{Sharpness-Aware Minimization} (SAM), which replaces pointwise loss minimization with a min-max objective that penalizes the worst-case loss within a perturbation ball of radius $\rho$. The inner maximization is approximated via a first-order Taylor expansion, yielding a two-step procedure that consistently improves generalization on CIFAR-10, CIFAR-100, and ImageNet~\cite{foret2020sharpness}. Subsequent extensions include ASAM~\cite{kwon2021asam}, which introduces adaptive sharpness invariant to parameter rescaling, and GSAM~\cite{zhuang2022surrogate}, which minimizes a surrogate gap related to the dominant Hessian eigenvalue. Despite these advances, all SAM variants optimize flatness at \emph{individual parameter vectors}; none extend sharpness-aware regularization to \emph{continuous paths} through weight space.

\subsection{Mode Connectivity and Weight-Space Geometry}\label{subsec:related_mc}
Garipov et al.~\cite{modeConnectivity} and Draxler et al.~\cite{draxler2018essentially} independently demonstrated that SGD-found minima are not isolated: independently trained solutions can be connected by simple parametric curves along which training loss and test error remain nearly constant. This discovery enabled \textit{Fast Geometric Ensembling} (FGE)~\cite{modeConnectivity} and \textit{Stochastic Weight Averaging} (SWA)~\cite{izmailov2018averaging}, both of which exploit the low-loss valley structure to improve generalization within a single training budget. Maddox et al.~\cite{maddox2019swag} extended SWA to Bayesian inference via SWAG. Ren et al.~\cite{ren2025revisiting} further extended conventional curve-based mode connectivity to B\'ezier surfaces.

The mode connectivity perspective has since intersected with model merging. Entezari et al.~\cite{entezari2022permutation} conjectured that accounting for permutation symmetry would eliminate loss barriers on linear interpolation paths; Ainsworth et al.~\cite{ainsworth2023git} validated this with \textit{Git Re-Basin}. Wortsman et al.~\cite{wortsman2022model} showed that simple weight averaging of fine-tuned models (\textit{Model Soups}) improves accuracy without increasing inference cost. Parameter-space connectivity has also been explored for machine unlearning~\cite{shi2025exploring} and evolutionary attacks~\cite{kim2026moco}. However, all of these methods optimize parameter-space structures using standard loss functions that do not explicitly account for the curvature of the surrounding loss surface.

\section{Proposed Method}\label{sec:method}

This section develops the Sharp Mode Connectivity (SMC) framework. We first review how Mode Connectivity constructs low-loss paths between independently trained solutions, then introduce Sharpness-Aware Minimization and the gap it addresses, and finally derive the SMC algorithm that integrates both perspectives.

\subsection{How to Build a Low-Loss Path: Mode Connectivity}\label{subsec:path_obj}

Garipov et al.~\cite{modeConnectivity} demonstrated that independently trained DNN solutions can be connected by simple parametric curves along which both training loss and test error remain nearly constant. Given a labeled dataset $\dataset = \{(x_i, y_i)\}_{i=1}^n$ drawn from an unknown distribution $\dataD$ over $\mathcal{X} \times \mathcal{Y}$, a DNN $f(\cdot\,; w) : \mathcal{X} \to \R^C$ with parameters $w \in \R^d$ incurs the empirical risk $\loss_\dataset(w) = \frac{1}{n} \sum_{i=1}^n \ell(f(x_i; w), y_i)$, where $\ell$ denotes the cross-entropy loss. Let $w_1, w_2 \in \R^d$ denote two independently trained solutions. Given two fixed endpoints, Mode Connectivity seeks a parametric curve $\phi_\theta : [0,1] \to \R^d$ satisfying the boundary conditions
\begin{equation}
	\phi_\theta(0) = w_1, \quad \phi_\theta(1) = w_2,
	\label{eq:boundary}
\end{equation}
where $\theta$ denotes learnable curve parameters (e.g., control points). The ideal objective minimizes the arc-length-normalized integral of the loss along the curve; however, since gradients of this objective are intractable, the practical formulation samples t from the uniform distribution $U(0,1)$:
\begin{equation}
	\loss_{\mathrm{path}}(\theta) = \E_{t \sim U(0,1)}\big[\loss_\dataset(\phi_\theta(t))\big].
	\label{eq:path_loss}
\end{equation}
We employ a \textbf{cubic B\'{e}zier curve} with two control points $\theta = (\theta_1, \theta_2)$ to parametrize the path:
\begin{equation}
	\phi_\theta(t) = (1-t)^3 w_1 + 3t(1-t)^2\theta_1 + 3t^2(1-t)\theta_2 + t^3 w_2.
	\label{eq:bezier_cubic}
\end{equation}

The choice of a cubic B\'{e}zier curve balances expressiveness and parsimony. Linear interpolation ($\phi(t) = (1-t)w_1 + tw_2$) provides no learnable degrees of freedom, so the path is fully determined by the endpoints and no optimization is possible. A quadratic B\'{e}zier curve introduces only one learnable control point, so the geometry of the early and late portions of the path cannot be adjusted independently. The cubic B\'{e}zier curve uses two control points and is therefore the lowest-order B\'{e}zier parameterization that provides separate control over the path near the two endpoints, which is useful for navigating asymmetric and non-convex loss landscapes.
Higher-order B\'{e}zier curves ($\textcolor{OliveGreen}{\geq 3}$ control points) would increase the dimensionality of the optimization problem in $\R^d$ without empirical evidence of improved path quality, and risk overfitting to noise in the sampled loss landscape.

To optimize the curve parameters $\theta$, we require the gradient of the path objective~\eqref{eq:path_loss}. This derivation proceeds in three steps. First, differentiating under the expectation yields
\begin{equation}
	\nabla_\theta \loss_{\mathrm{path}}(\theta)
	= \E_{t \sim U(0,1)}\!\Big[\nabla_\theta \loss_\dataset\!\big(\phi_\theta(t)\big)\Big].
	\label{eq:grad_step1}
\end{equation}
Since the loss $\loss_\dataset$ is defined as a function of the weights $w = \phi_\theta(t) \in \R^d$ and the curve $\phi_\theta$ maps the parameters $\theta$ into weight space, we apply the chain rule to decompose the gradient:
\begin{equation}
	\nabla_\theta \loss_\dataset\!\big(\phi_\theta(t)\big)
	= \underbrace{\left(\frac{\partial \phi_\theta(t)}{\partial \theta}\right)^{\!\top}}_{\text{curve Jacobian}} \underbrace{\nabla_w \loss_\dataset\!\big(\phi_\theta(t)\big)}_{\text{loss gradient in weight space}}.
	\label{eq:grad_step2}
\end{equation}
Here $\nabla_w \loss_\dataset(\phi_\theta(t))$ is the standard loss gradient evaluated at the weight vector $\phi_\theta(t)$, and $\frac{\partial \phi_\theta(t)}{\partial \theta}$ is the Jacobian of the B\'{e}zier parameterization~\eqref{eq:bezier_cubic} with respect to the control points, whose entries are the basis polynomials $3t(1-t)^2$ and $3t^2(1-t)$ evaluated at~$t$. Substituting~\eqref{eq:grad_step2} into~\eqref{eq:grad_step1} gives the complete path gradient:
\begin{equation}
    \nabla_\theta \loss_{\mathrm{path}}(\theta) =\mathbb{E}_{t\sim U(0,1)}\left[\left(\frac{\partial \phi_{\theta}(t)}{\partial \theta}\right)^{\top}\nabla_{w}\loss_\dataset\big(\phi_{\theta}(t)\big)\right].
    \label{eq:path_grad}\nonumber
\end{equation}

\subsection{How to Find Flat Minima: Sharpness-Aware Minimization}\label{subsec:sam_obj}

The geometry of loss landscape minima strongly influences generalization: sharp minima, where the loss increases rapidly under small perturbations, tend to generalize poorly, whereas flat minima remain low-loss throughout a local neighborhood~\cite{foret2020sharpness}. Sharpness-Aware Minimization (SAM) formalizes this insight by seeking parameters whose entire $\ell_2$-ball of radius $\rho > 0$ attains low loss. The \emph{sharpness-aware loss} is defined as
\begin{equation}
	\loss_\dataset^{\mathrm{SAM}}(w) \triangleq \max_{\|\epsilon\|_2 \leq \rho} \loss_\dataset(w + \epsilon),
	\label{eq:sam_loss}
\end{equation}
where $\epsilon \in \R^d$ is a perturbation vector and $\rho$ controls the neighborhood size. Since solving the inner maximization exactly is intractable, a first-order Taylor expansion yields the closed-form approximation for the optimal perturbation:
\begin{equation}
	\hat{\epsilon}(w) = \rho \frac{\nabla_w \loss_\dataset(w)}{\|\nabla_w \loss_\dataset(w)\|_2}.
	\label{eq:sam_epsilon}
\end{equation}
The SAM gradient is then approximated as $\nabla_w \loss_\dataset(w + \hat{\epsilon}(w))$, dropping second-order terms for computational efficiency. This results in a two-step procedure at each training iteration: first, compute the ascent perturbation using Equation~\eqref{eq:sam_epsilon}, then evaluate the gradient at the perturbed point.

These two formulations are complementary but have been studied in isolation: MC discovers low-loss curves~\cite{modeConnectivity, draxler2018essentially} but imposes no constraint on sharpness along them, while SAM~\cite{foret2020sharpness} enforces flatness within a local neighborhood of individual parameter vectors without extending to continuous trajectories. To our knowledge, no existing method integrates sharpness constraints into the path-finding objective.

This gap motivates the central question of this paper: \emph{can we find a connectivity path between independently trained solutions that is not only low-loss but also uniformly flat within a local neighborhood along its entire trajectory?} We answer this question affirmatively in the following subsection.

\subsection{How to Find the Flat Path: Sharp Mode Connectivity}\label{subsec:smc_algorithm}

\subsubsection{Sharpness-Aware Path Functional}\label{subsec:hybrid}

Instead of evaluating the standard empirical loss $\loss_\dataset(\phi_\theta(t))$ at each point along the curve, we evaluate the worst-case loss within a $\rho$-neighborhood. The \emph{sharpness-aware path functional} is defined as
\begin{equation}
    \min_{\theta}
	J_{\mathrm{sharp}}(\theta) = \int_0^1 \max_{\|\epsilon\|_2 \leq \rho} \loss_\dataset\big(\phi_\theta(t) + \epsilon\big)\,dt.
	\label{eq:sharp_path}
\end{equation}
This objective encourages points along the path, on average over $t$, to lie in neighborhoods with low worst-case loss, extending SAM's local minimax reasoning from individual parameter vectors to a continuous connectivity trajectory.
Minimizing~\eqref{eq:sharp_path} encourages the optimizer to discover wide, smooth valleys connecting the modes $w_1$ and $w_2$, rather than paths that merely achieve low loss at each point but may traverse sharp ridges vulnerable to distributional perturbation.

\subsubsection{SAM-on-the-Curve Approximation}\label{subsec:sam_on_curve}

Direct evaluation of the inner maximization in~\eqref{eq:sharp_path} is intractable. Applying the first-order SAM approximation at each curve point, the perturbation is
\begin{equation}
	\hat{\epsilon}(\phi_\theta(t)) = \rho \frac{\nabla_w \loss_\dataset(\phi_\theta(t))}{\|\nabla_w \loss_\dataset(\phi_\theta(t))\|_2}.
	\label{eq:curve_epsilon}
\end{equation}
Two approximations make this objective tractable. First, the inner maximization at each curve point is replaced by the standard first-order SAM perturbation~\eqref{eq:curve_epsilon}, which computes the steepest ascent direction within the $\rho$-ball. Second, the continuous integral over $t \in [0,1]$ is estimated via Monte Carlo sampling: $K$ points $\{t_k\}_{k=1}^K$ are drawn uniformly from $[0,1]$ and the functional is approximated by their average. These two approximations yield the practical objective:
\begin{equation}
	J_{\mathrm{sharp}}(\theta) \approx \frac{1}{K} \sum_{k=1}^K \loss_\dataset\Big(\phi_\theta(t_k) + \hat{\epsilon}\big(\phi_\theta(t_k)\big)\Big).
	\label{eq:sharp_mc}
\end{equation}

\subsubsection{Gradient with Respect to Curve Parameters}\label{subsec:sharp_grad}

Let $w_t = \phi_\theta(t)$ and define the SAM gradient at the perturbed point as $g_{\mathrm{SAM}}(w_t) = \nabla_w \loss_\dataset(w_t + \hat{\epsilon}(w_t))$. By the chain rule:
\begin{equation}
	\nabla_\theta J_{\mathrm{sharp}}(\theta) \approx \E_{t \sim U(0,1)}\left[\left(\frac{\partial \phi_\theta(t)}{\partial \theta}\right)^{\!\top} g_{\mathrm{SAM}}(w_t)\right],
	\label{eq:sharp_grad}
\end{equation}
where derivatives through $\hat{\epsilon}(w_t)$ are omitted for computational efficiency, following standard SAM practice. Each update of $\theta$ requires: (i)~computing $\nabla_w \loss_\dataset(w_t)$, (ii)~constructing $\hat{\epsilon}(w_t)$, (iii)~evaluating the loss gradient at the perturbed point, and (iv)~multiplying by the curve Jacobian $\partial \phi_\theta(t) / \partial \theta$. This procedure is computationally analogous to SAM, requiring two backpropagations per sampled~$t$, but operates on a curve in weight space rather than on a single parameter vector.

\subsubsection{Complete Algorithm}\label{subsec:algorithm}

Algorithm~\ref{alg:sharp_curve} implements the SMC framework as a practical optimization procedure. The goal is to find curve parameters $\theta$ that yield a low-loss connectivity path whose local neighborhood is uniformly flat, since flat regions in the loss landscape are associated with better generalization~\cite{foret2020sharpness}.

\begin{algorithm}[htbp]
	\caption{Sharp Mode Connectivity (SMC)}\label{alg:sharp_curve}
	\begin{algorithmic}[0.75]
		\REQUIRE Endpoints $w_1, w_2$, perturbation radius $\rho$, learning rate $\eta$, samples per step $K$, stabilizer $\delta$
		\STATE Initialize curve parameters $\theta = (\theta_1, \theta_2)$ \hfill\COMMENT{Cubic B\'{e}zier control points}
		\WHILE{not converged}
		\STATE Sample mini-batch $\mathcal{B}$
		\STATE Sample $\{t_k\}_{k=1}^K \sim U(0,1)$
		\STATE $G_\theta \leftarrow 0$
		\FOR{$k = 1$ to $K$}
		\STATE $w_k \leftarrow \phi_\theta(t_k)$ \hfill\COMMENT{Interpolate via~\eqref{eq:bezier_cubic}}
		\STATE $g_k \leftarrow \nabla_w \loss_\mathcal{B}(w_k)$ \hfill\COMMENT{Ascent direction}
		\STATE $\hat{\epsilon}_k \leftarrow \rho\, g_k / (\|g_k\|_2 + \delta)$ \hfill\COMMENT{SAM perturbation}
		\STATE $\tilde{g}_k \leftarrow \nabla_w \loss_\mathcal{B}(w_k + \hat{\epsilon}_k)$ \hfill\COMMENT{Descent gradient}
		\STATE $G_\theta \leftarrow G_\theta + \left(\partial\phi_\theta(t_k)/\partial\theta\right)^{\!\top} \tilde{g}_k$
		\ENDFOR
		\STATE $G_\theta \leftarrow G_\theta / K$
		\STATE $\theta \leftarrow \theta - \eta\, G_\theta$
		\ENDWHILE
		\RETURN Optimized curve parameters $\theta$
	\end{algorithmic}
\end{algorithm}

The algorithm uses a single mini-batch $\mathcal{B}$ for both the ascent (perturbation) and descent (update) steps, ensuring consistency between the two gradient computations and reducing variance from batch-level stochasticity. The stabilizer $\delta$ prevents division by zero in the perturbation normalization.

\section{Experimental Results}\label{sec:experiments}

\subsection{Models and Datasets}\label{subsec:arch}
We evaluate the proposed framework across three architectures spanning distinct design families:
\begin{itemize}
	\item \textbf{ResNet-18}~\cite{he2016resnet}: Modified for CIFAR-10 with a $3\times3$ initial convolution (stride 1, padding 1, no bias) replacing the standard $7\times7$ convolution, and an identity mapping replacing the max-pooling layer to accommodate the $32\times32$ spatial dimensions. Serves as the primary architecture for all ablation studies.
	\item \textbf{VGG16-BN}~\cite{simonyan2015vgg}: Standard VGG16 with Batch Normalization, providing a purely sequential architecture without residual connections. Tests whether the framework generalizes beyond residual networks.
	\item \textbf{ViT-Tiny}~\cite{dosovitskiy2021vit}: A compact Vision Transformer with patch embedding and multi-head self-attention, testing attention-based architectures.
\end{itemize}
Experiments are conducted on two datasets:
\begin{itemize}
	\item \textbf{CIFAR-10}: 60,000 images in 10 classes ($32\times32$). Used for all three architectures and all ablation studies.
	\item \textbf{ImageNet-100}: A 100-class subset of ImageNet ($224\times224$). Used with ViT-Tiny to test Mode Connectivity on higher-resolution data with attention-based architectures.
\end{itemize}
Out-of-distribution evaluation uses the \textbf{CIFAR-10-C} benchmark~\cite{hendrycks2019robustness, hendrycks2019zenodo_cifar10c}, comprising 15 corruption types at 5 severity levels. We focus on the ``Blur'' category (Gaussian, Defocus, Glass, and Motion blur) as these systematically degrade spatial frequency information.

\subsection{Implementation Details}\label{subsec:impl}

Batch Normalization statistics require special handling: interpolated running statistics are mathematically invalid for non-linear networks. We perform \textbf{recalibration} by running 100 gradient-free forward passes on clean training data at each evaluation point $t$ to recompute valid running means and variances. This recalibration uses only clean data to prevent target distribution leakage. The hyperparameters are summarized in Table~\ref{tab:hyperparams}. All experiments use fixed random seeds to ensure full reproducibility.

\begin{table}[htbp]
	\centering
	\caption{Hyperparameters for Sharp Mode Connectivity}\label{tab:hyperparams}
	\renewcommand{\arraystretch}{1.1}
	\begin{tabular}{lc}
		\toprule
		\textbf{Parameter} & \textbf{Value} \\
		\midrule
		Architecture & Modified ResNet-18 \\
		Dataset & CIFAR-10 \\
		Curve parametrization & Cubic B\'{e}zier \\
		Batch size ($|\mathcal{B}|$) & 128 \\
		SAM neighborhood radius ($\rho$) & 0.05 \\
		Interpolation samples ($K$) & 4 \\
		Pre-training epochs ($w_1, w_2$) & 30 \\
		Curve training epochs & 50 \\
		Pre-training learning rate & 0.1 \\
		Curve learning rate ($\eta$) & 0.01 \\
		Optimizer & SGD (momentum = 0.9) \\
		Weight decay & $1\times10^{-4}$ (curve), $5\times10^{-4}$ (pre-training) \\
		LR scheduler & Cosine annealing \\
		\bottomrule
	\end{tabular}
\end{table}

\subsection{Evaluation Protocol}\label{subsec:eval_protocol}

\subsubsection{Peak vs. Midpoint Accuracy}
We report two accuracy metrics along the path: \emph{peak accuracy} ($\max_{t} \text{Acc}(t)$) to identify the single best model for ensembling and model soups, and \emph{midpoint accuracy} ($t=0.5$) to assess connectivity robustness, as the midpoint is furthest from the endpoints and most vulnerable to barriers and distribution shifts.

\subsubsection{Midpoint Loss Barrier} We define the midpoint loss barrier as:
\begin{equation}
	\Delta\loss_{\mathrm{barrier}} = \loss\big(\phi_\theta(0.5)\big) - \frac{1}{2}\Big(\loss(w_1) + \loss(w_2)\Big).
	\label{eq:loss_barrier}
\end{equation}
A value of zero indicates perfect barrier-free interpolation. Positive values indicate sharp ridges; negative values indicate that the midpoint loss is lower than the average endpoint loss.

\subsubsection{Point-wise Sharpness (SAM Gap)} We measure the sharpness of the loss surface at each curve point $\phi_\theta(t)$ using the SAM Gap:
\begin{equation}
	\Delta_{\mathrm{SAM}}(t) = \max_{\|\epsilon\|_2 \leq \rho} \loss_\dataset\big(\phi_\theta(t) + \epsilon\big) - \loss_\dataset\big(\phi_\theta(t)\big).
	\label{eq:sam_gap}
\end{equation}
This quantity measures how much the loss increases under worst-case perturbation at position~$t$ along the curve. A high SAM Gap indicates a sharp ridge; a low SAM Gap indicates a flat valley. We report both the path-averaged SAM Gap $\bar{\Delta}_{\mathrm{SAM}} = \frac{1}{|T|}\sum_{t \in T} \Delta_{\mathrm{SAM}}(t)$ and the midpoint value $\Delta_{\mathrm{SAM}}(0.5)$.

\subsection{Sanity Checks}\label{subsec:sanity}

\begin{figure}[htbp]
	\centering
	\includegraphics[width=0.25\textwidth]{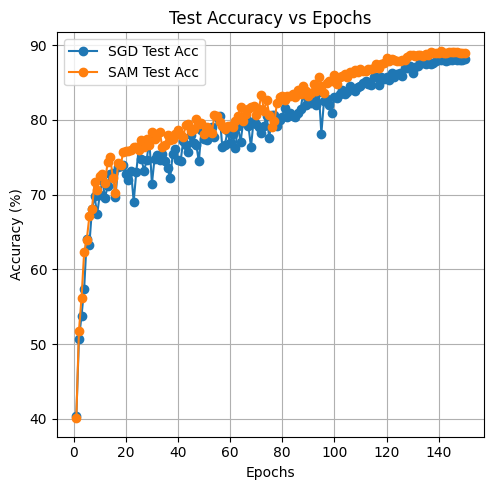}
	\caption{SGD vs.\ SAM training metrics on ResNet-18/CIFAR-10. SAM achieves higher test accuracy, validating its sharpness-aware optimization.}
	\label{fig:sam_baseline}
\end{figure}

\textbf{SAM optimizer.} We verify the standalone SAM optimizer by comparing it against baseline SGD on ResNet-18/CIFAR-10 (150 epochs, batch size 128, cosine annealing, $\rho=0.05$). As shown in Fig.~\ref{fig:sam_baseline}, SAM consistently outperforms SGD in test accuracy while both achieve near-89\%  training accuracy, confirming that SAM finds flatter, better-generalizing minima.

\begin{figure}[htbp]
	\centering
	\begin{subfigure}[b]{0.235\textwidth}
		\centering
		\includegraphics[width=\textwidth]{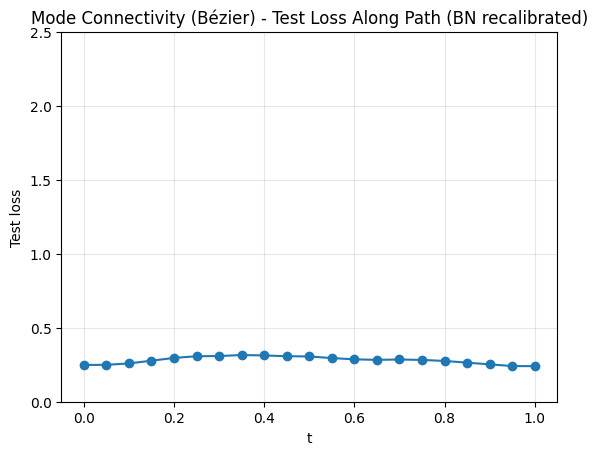}
		\caption{Test loss along path}
		\label{fig:mc_loss}
	\end{subfigure}
	\hfill
	\begin{subfigure}[b]{0.235\textwidth}
		\centering
		\includegraphics[width=\textwidth]{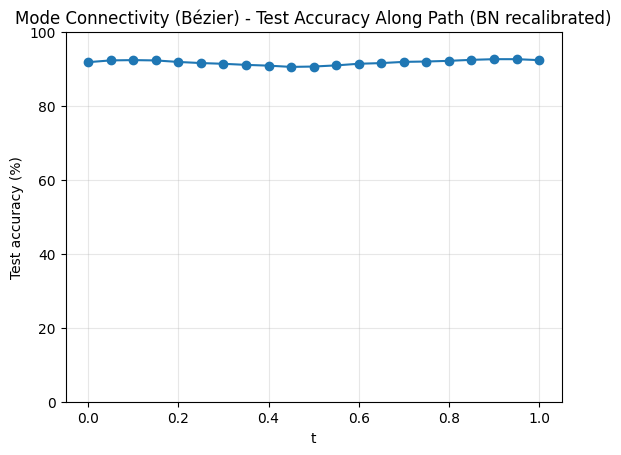}
		\caption{Test accuracy along path}
		\label{fig:mc_acc}
	\end{subfigure}
	\caption{Mode Connectivity verification on ResNet-18/CIFAR-10: test loss (a) and accuracy (b) along the optimized cubic B\'{e}zier path after BN recalibration. The path maintains $>$90\% accuracy across the entire trajectory, confirming a low-loss, barrier-free connection between independently trained solutions.}
	\label{fig:mc_curves}
\end{figure}

\textbf{Mode Connectivity.}
Two independent ResNet-18 models ($w_1$: 91.68\%, $w_2$: 92.44\%) are connected via an optimized cubic B\'{e}zier curve. After Batch Normalization recalibration, the path exhibits uniformly high performance: test accuracy remains above 90\% across the entire trajectory (Fig.~\ref{fig:mc_curves}), confirming the existence of a smooth, low-loss connection between independently trained solutions.

\subsection{Clean Connectivity: Standard MC vs.\ SMC}\label{subsec:clean_conn}

We first compare the standard MC path against the SMC path on clean test data across three architectures: ResNet-18, VGG16-BN, and ViT-Tiny.

\subsubsection{CNN Architectures (ResNet-18 and VGG16-BN)}

\begin{figure}[htbp]
	\centering
	\includegraphics[width=0.5\textwidth]{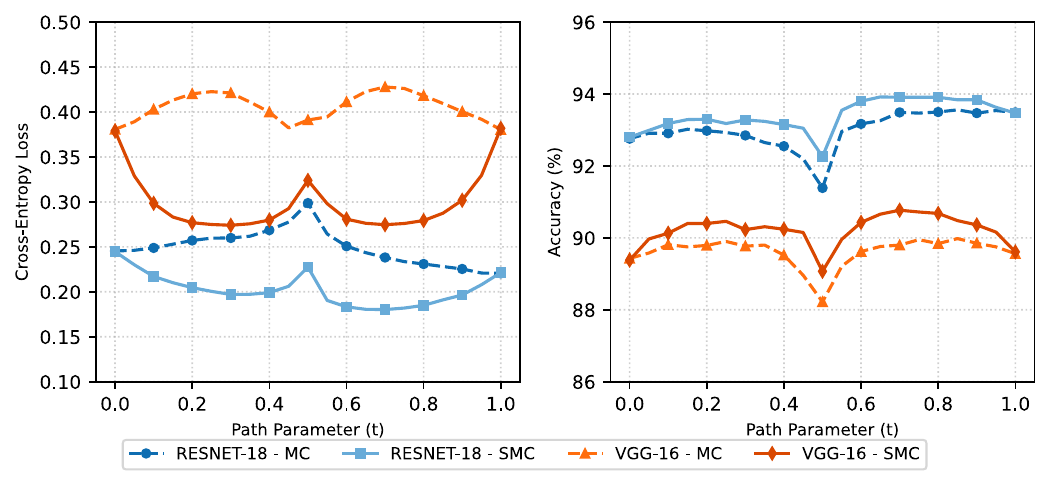}
	\caption{Clean connectivity landscape on CIFAR-10. \textbf{Left}: Cross-entropy loss vs.\ path parameter $t$ for the SMC model (blue) and standard MC (red) across both ResNet-18 and VGG16-BN architectures. \textbf{Right}: Test accuracy (\%). The SMC maintains a flat ``U-shape'' loss profile and higher accuracy across the full trajectory, confirming that the benefit generalizes beyond residual architectures.}
	\label{fig:clean_connectivity}
\end{figure}

Fig.~\ref{fig:clean_connectivity} presents the clean connectivity landscape for both CNN architectures.

\begin{figure}[htbp]
	\centering
	\includegraphics[width=0.35\textwidth]{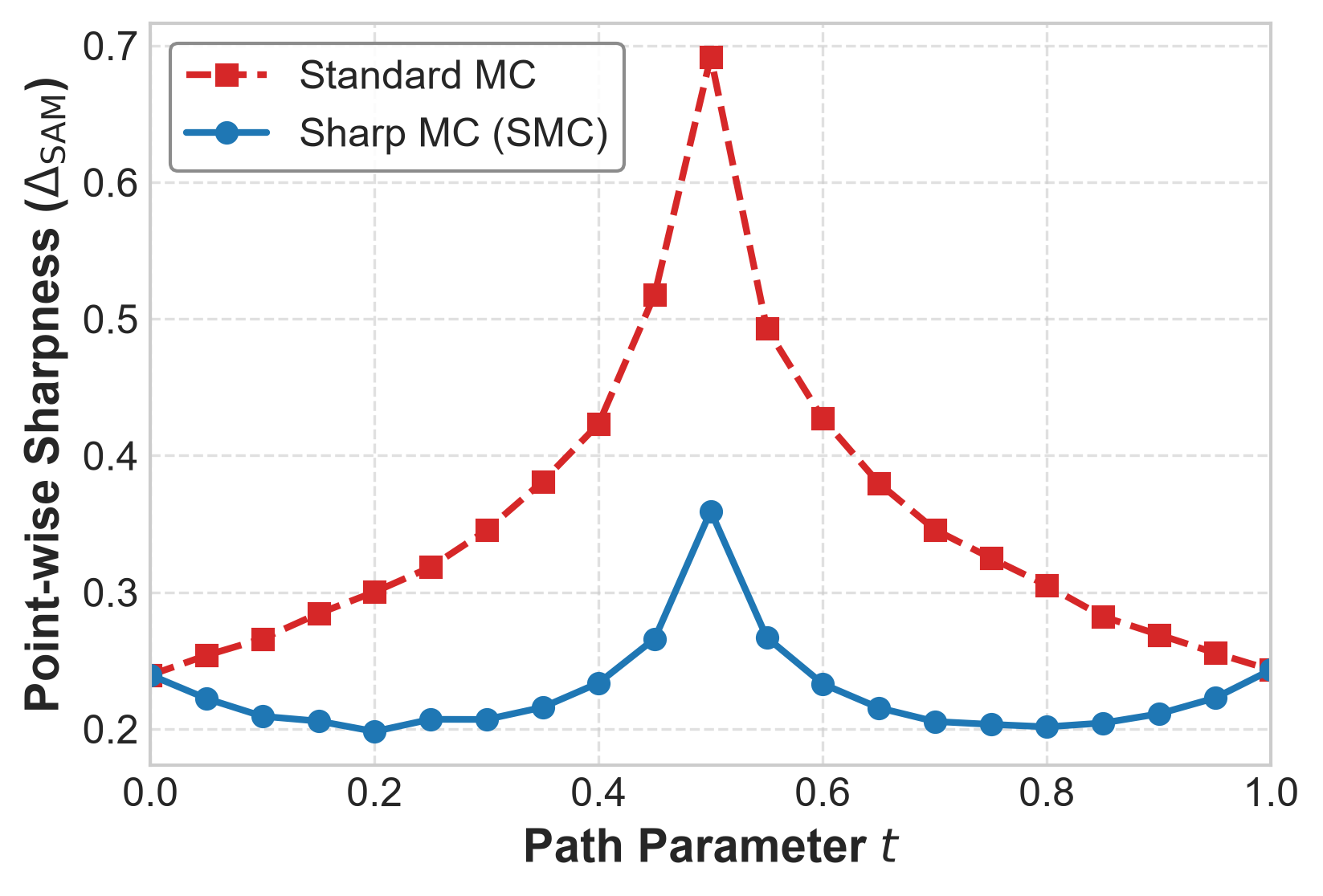}
	\caption{Point-wise sharpness (SAM Gap) along the connectivity path for ResNet-18 on CIFAR-10. The standard MC path (red, dashed) exhibits a sharp spike at the midpoint ($\Delta_{\mathrm{SAM}}(0.5) = 0.6916$), revealing that the curve threads through a narrow ridge at $t=0.5$. The SMC path (blue, solid) suppresses sharpness across the entire trajectory, peaking at only $0.3594$ at the midpoint, a $48\%$ reduction. Both paths share identical endpoints ($t=0$ and $t=1$), confirming that the sharpness reduction arises entirely from the path-interior optimization.}
	\label{fig:sam_gap_profile}
\end{figure}

Fig.~\ref{fig:sam_gap_profile} plots the point-wise sharpness (SAM Gap, Eq.~\ref{eq:sam_gap}) at each curve location for both methods on ResNet-18.

\textbf{ResNet-18 (CIFAR-10)}: The standard MC trajectory exhibits a distinct ``V-shape'' with a loss peak of 0.2851 at the midpoint. In contrast, the SMC model maintains a remarkably stable, deep ``U-shape'' with substantially lower loss throughout. The standard MC model drops to 92.07\% accuracy at the midpoint, while the SMC maintains 92.33\%. Notably, both frameworks discover a region of superior generalization at $t=0.90$, where the SMC achieves 94.00\%, surpassing both independently trained endpoints.

\textbf{VGG16-BN (CIFAR-10)}: As shown inFig.~\ref{fig:clean_connectivity}, SMC achieves a flatter loss profile and higher accuracy along the entire path. The SMC attains a path-averaged accuracy of $90.22\%$ vs.\ $89.61\%$ for standard MC, with path loss reduced from 0.406 to 0.299, a $26.4\%$ reduction. Despite lacking skip connections that smooth the loss landscape, VGG16-BN exhibits clear mode connectivity, confirming that the SAM-based framework generalizes beyond residual architectures.

\begin{mydef}[Path Loss Flatness]\label{def:flatness}
We define the \emph{path loss flatness} as the variance of the loss evaluated at uniformly spaced points along the curve:
\begin{equation}
	F(\theta) = \operatorname{Var}_{t \in T}\big[\loss_\dataset(\phi_\theta(t))\big] = \frac{1}{|T|}\sum_{t \in T}\Big(\loss_\dataset(\phi_\theta(t)) - \bar{\loss}\Big)^2,
	\label{eq:flatness}
\end{equation}
where $T$ is a set of uniformly spaced evaluation points along the curve and $\bar{\loss} = \frac{1}{|T|}\sum_{t \in T}\loss_\dataset(\phi_\theta(t))$ is the path-averaged loss. Lower values indicate a more uniform loss profile along the path: the curve traverses a consistently deep valley rather than oscillating between high- and low-loss regions.
\end{mydef}

Table~\ref{tab:clean_baseline} summarizes the clean baseline performance across all three architectures. The loss flatness metric (Definition~\ref{def:flatness}) quantifies the uniformity of the loss profile along the path. For ResNet-18, the SMC achieves a $26.7\times$ improvement in flatness (0.0021 vs.\ 0.056), confirming that SAM's min-max regularization successfully enforces uniform curvature constraints across the connectivity manifold.

The SAM Gap profile (Fig.~\ref{fig:sam_gap_profile}) reveals the geometric mechanism behind this improvement. Along the standard MC path, point-wise sharpness rises sharply toward the midpoint, peaking at $\Delta_{\mathrm{SAM}}(0.5) = 0.6916$, indicating that the path traverses an increasingly sharp ridge as it departs from either endpoint. This is precisely the location where accuracy degrades most under distribution shift (Section~\ref{subsec:ood}). The SMC path maintains substantially lower sharpness throughout, with the midpoint SAM Gap reduced to $0.3594$ (a $48\%$ reduction) and the path-averaged SAM Gap reduced from $0.3499$ to $0.2271$ (a $35\%$ reduction). This provides direct empirical evidence that SMC's sharpness-aware objective flattens the loss surface not only on the curve but in its transverse neighborhood, establishing the causal link between path-wise flatness and the OOD robustness gains reported in Section~\ref{subsec:ood}.

\begin{table}[htbp]
	\centering
    	\caption{Clean Baseline Performance Metrics}\label{tab:clean_baseline}
	\renewcommand{\arraystretch}{1.1}
	\begin{tabular}{llcc}
		\toprule
		\textbf{Architecture} & \textbf{Metric} & \textbf{SMC} & \textbf{MC} \\
		\midrule
		\multirow{6}{*}{ResNet-18}
		& Midpoint accuracy (\%) & 92.33 & 92.07 \\
		& Path accuracy (\%) & 93.521 & 93.225 \\
		& Path CE loss & 0.197 & 0.244 \\
		& Loss flatness (var.) & 0.0021 & 0.056 \\
		& Avg SAM Gap & 0.2271 & 0.3499 \\
		& Midpoint SAM Gap & 0.3594 & 0.6916 \\
		\midrule
		\multirow{2}{*}{VGG16-BN}
		& Path accuracy (\%) & 90.218 & 89.612 \\
		& Path CE loss & 0.299 & 0.406 \\
		\midrule
		\multirow{4}{*}{\shortstack[l]{ViT-Tiny\\(CIFAR-10)}}
		& Midpoint accuracy (\%) & 77.30 & 76.40 \\
		& Best path accuracy (\%) & 87.63 & 87.55 \\
		& Path accuracy (\%) & 86.08 & 85.84 \\
		& Path CE loss & 0.485 & 0.541 \\
		\midrule
		\multirow{4}{*}{\shortstack[l]{ViT-Tiny\\(ImageNet-100)}}
		& Midpoint accuracy (\%) & 68.08 & 67.53 \\
		& Best path accuracy (\%) & 75.65 & 74.18 \\
		& Path accuracy (\%) & 74.321 & 73.316 \\
		& Path CE loss & 1.003 & 1.120 \\
		\bottomrule
	\end{tabular}
\end{table}

\subsubsection{Vision Transformer (ViT-Tiny)}

\begin{figure}[htbp]
    \centering
    \includegraphics[width=0.95\linewidth]{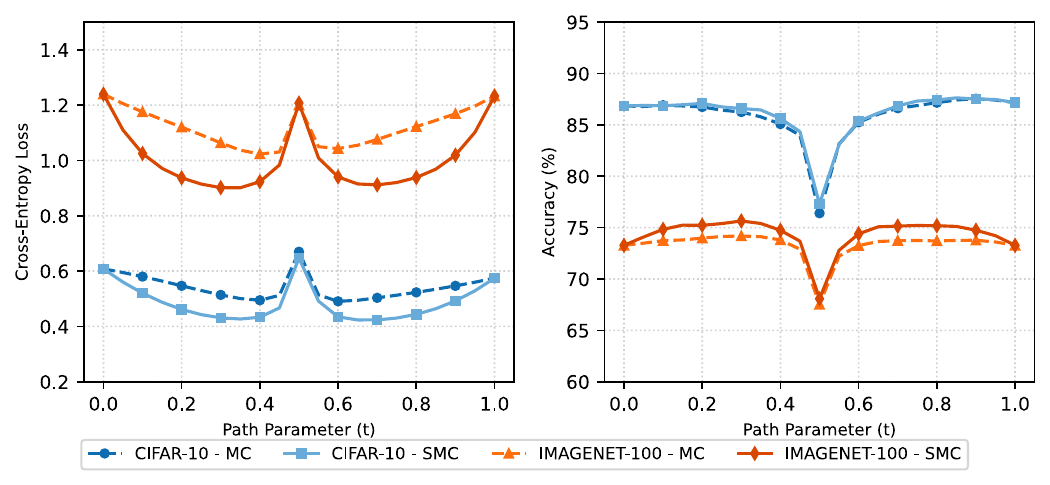}
    \caption{ViT-Tiny clean connectivity on CIFAR-10 (blue) and ImageNet-100 (red). Solid lines denote the SMC path; dashed lines denote standard MC. \textbf{Left}: Cross-entropy loss along the path parameter~$t$. \textbf{Right}: Test accuracy (\%). The SMC achieves lower loss and higher accuracy along the path interior on both datasets.}\label{fig:vit_clean}
\end{figure}

The ViT-Tiny results (Fig.~\ref{fig:vit_clean}) reveal a distinct connectivity structure compared to CNNs. While both architectures exhibit low-loss connectivity away from the midpoint, ViT-Tiny displays a residual midpoint barrier: accuracy dips to $\sim$77\% on CIFAR-10 and $\sim$68\% on ImageNet-100 at $t=0.5$, compared to the $\sim$87\% and $\sim$73\% at the endpoints, respectively. Crucially, however, the SMC model consistently outperforms standard MC across both datasets. On CIFAR-10, the SMC achieves a path-averaged accuracy of 86.08\% vs.\ 85.84\% for standard MC, with lower path loss (0.485 vs.\ 0.541). On ImageNet-100, the advantage is more pronounced: 74.32\% vs.\ 73.32\% path accuracy (1.003 vs.\ 1.120 path loss). The SMC also discovers points along the path that surpass both endpoints, reaching 87.63\% at $t=0.85$ on CIFAR-10 and 75.65\% at $t=0.30$ on ImageNet-100, confirming that the sharpness-aware path functional identifies superior solutions even in attention-based architectures.

These results reframe the ViT connectivity landscape: rather than a failure of the framework, the residual midpoint barrier reflects an architectural property of attention mechanisms, while the consistent SMC advantage demonstrates that sharpness-aware path optimization remains effective across architecture families. We proceed to stress-test the CNN architectures under severe distribution shifts.

\subsection{Out-of-Distribution Robustness (CIFAR-10-C)}\label{subsec:ood}

We evaluate the SMC framework under severe distribution shifts using blur corruptions from CIFAR-10-C at five severity levels, comparing ResNet-18 and VGG16-BN.

\subsubsection{Accuracy Under Distribution Shift}

\begin{figure}[htbp]
	\centering
	\includegraphics[width=0.5\textwidth]{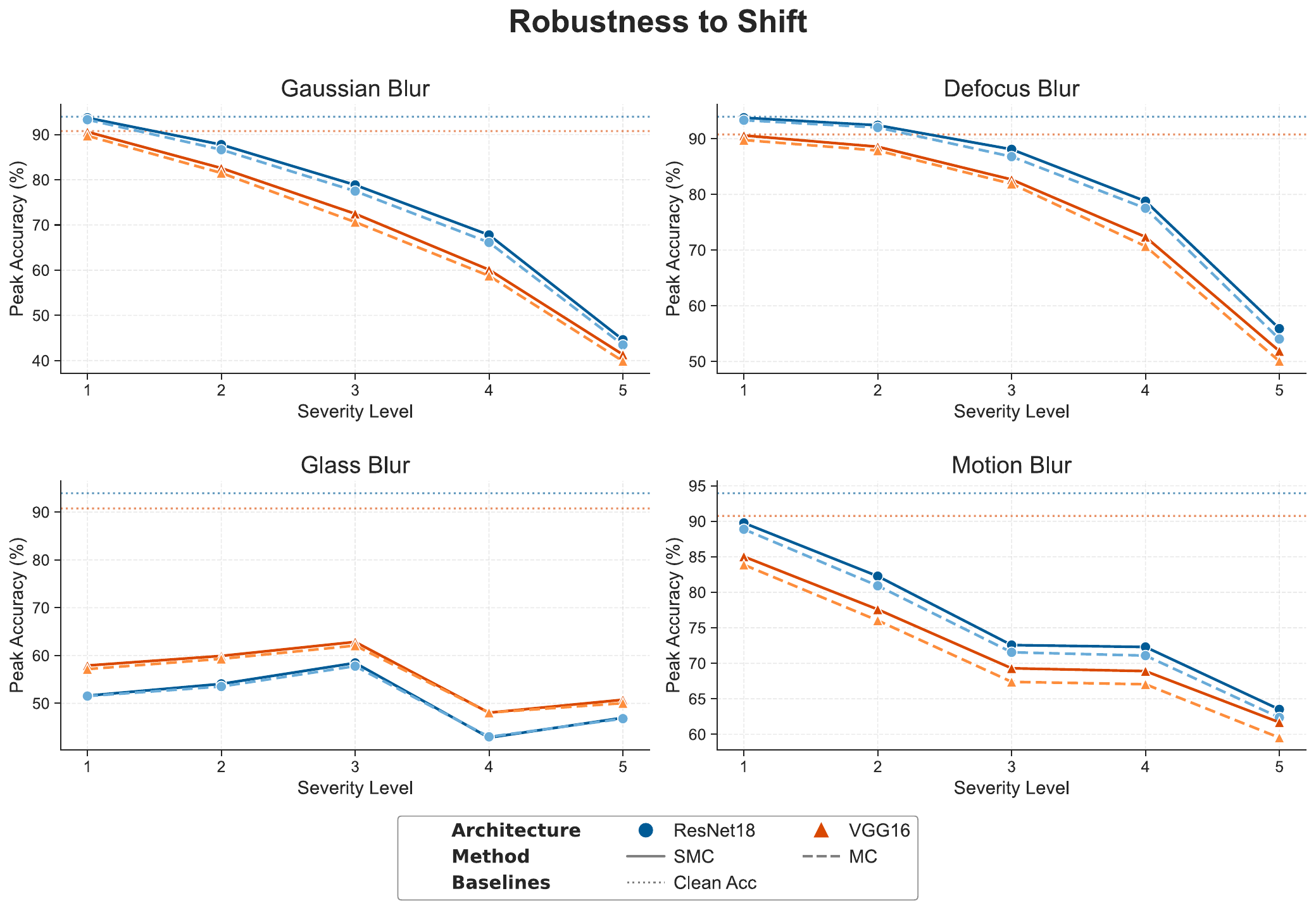}
	\caption{ResNet-18 midpoint accuracy across blur corruption severities for Gaussian, Defocus, Glass, and Motion blur on CIFAR-10-C. The SMC consistently outperforms standard MC, with the advantage generally scaling with severity. The clean baseline is marked with a dashed gray line.}
	\label{fig:midpoint_acc}
\end{figure}

Fig.~\ref{fig:midpoint_acc} plots the accuracy across severity levels for all four blur types. The SMC model consistently outperforms standard MC, with the advantage generally growing with corruption severity. Under Gaussian blur (Severity 5), the SMC maintains 35.98\% vs.\ 30.27\% for standard MC; under Defocus blur, 52.44\% vs.\ 46.35\%.

Table~\ref{tab:sev5_acc} quantifies the Severity 5 advantage across both CNN architectures. The maximum absolute improvement is $+6.09\%$ under Defocus blur for ResNet-18 and $+4.08\%$ for VGG16-BN.

\begin{table}[htbp]
	\centering
	\caption{Midpoint Accuracy Under Severe Shift (Severity 5): ResNet-18 and VGG16-BN}\label{tab:sev5_acc}
	\renewcommand{\arraystretch}{1}
	\begin{tabular}{lcccccc}
		\toprule
		& \multicolumn{3}{c}{\textbf{ResNet-18}} & \multicolumn{3}{c}{\textbf{VGG16-BN}} \\
		\cmidrule(lr){2-4} \cmidrule(lr){5-7}
		\textbf{Corruption} & \textbf{SMC} & \textbf{MC} & \textbf{$\Delta$} & \textbf{SMC} & \textbf{MC} & \textbf{$\Delta$} \\
		\midrule
		Gaussian Blur & 35.98 & 30.27 & +5.71 & 39.88 & 36.09 & +3.79 \\
		Defocus Blur & 52.44 & 46.35 & +6.09 & 49.88 & 45.80 & +4.08 \\
		Glass Blur & 41.00 & 40.28 & +0.72 & 46.01 & 44.86 & +1.15 \\
		Motion Blur & 58.79 & 55.38 & +3.41 & 57.82 & 54.33 & +3.49 \\
		\bottomrule
	\end{tabular}
\end{table}

Table~\ref{tab:adv_grid} reports the SMC accuracy advantage across all severity levels for both architectures, revealing a broadly monotonic scaling trend: as corruption intensifies, the advantage of flat-basin connectivity generally increases. The trend is strongest for Gaussian, Defocus, and Motion blur, though Glass blur deviates with non-monotonic reversals. Notably, the VGG16-BN SMC consistently outperforms standard MC from Severity~1 onward, whereas for ResNet-18 the advantage emerges at Severity~2, suggesting that purely sequential architectures are more uniformly sensitive to sharpness-aware regularization.

\begin{table}[htbp]
	\centering
	\caption{Midpoint SMC Accuracy Advantage ($+\%$) Over Standard MC Across Severities}\label{tab:adv_grid}
	\renewcommand{\arraystretch}{1.1}
	\begin{tabular}{llcccc}
		\toprule
		\textbf{Model} & \textbf{Sev.} & \textbf{Gauss.} & \textbf{Defoc.} & \textbf{Glass} & \textbf{Mot.} \\
		\midrule
		\multirow{5}{*}{ResNet-18}
		& 1 & $-0.12$ & $-0.12$ & $-0.13$ & $+1.31$ \\
		& 2 & $+1.62$ & $+0.30$ & $-0.15$ & $+2.15$ \\
		& 3 & $+3.42$ & $+1.44$ & $+0.94$ & $+2.92$ \\
		& 4 & $+5.39$ & $+3.19$ & $-0.66$ & $+3.50$ \\
		& 5 & $+5.71$ & $+6.09$ & $+0.72$ & $+3.41$ \\
		\midrule
		\multirow{5}{*}{VGG16-BN}
		& 1 & $+1.26$ & $+1.30$ & $+2.26$ & $+2.34$ \\
		& 2 & $+1.99$ & $+1.88$ & $+2.30$ & $+2.98$ \\
		& 3 & $+3.71$ & $+2.04$ & $+1.50$ & $+4.12$ \\
		& 4 & $+4.11$ & $+2.94$ & $+2.68$ & $+3.95$ \\
		& 5 & $+3.79$ & $+4.08$ & $+1.15$ & $+3.49$ \\
		\bottomrule
	\end{tabular}
\end{table}

\FloatBarrier
\subsubsection{Loss Barrier Analysis}

\begin{figure}[htbp]
	\centering
	\includegraphics[width=0.5\textwidth]{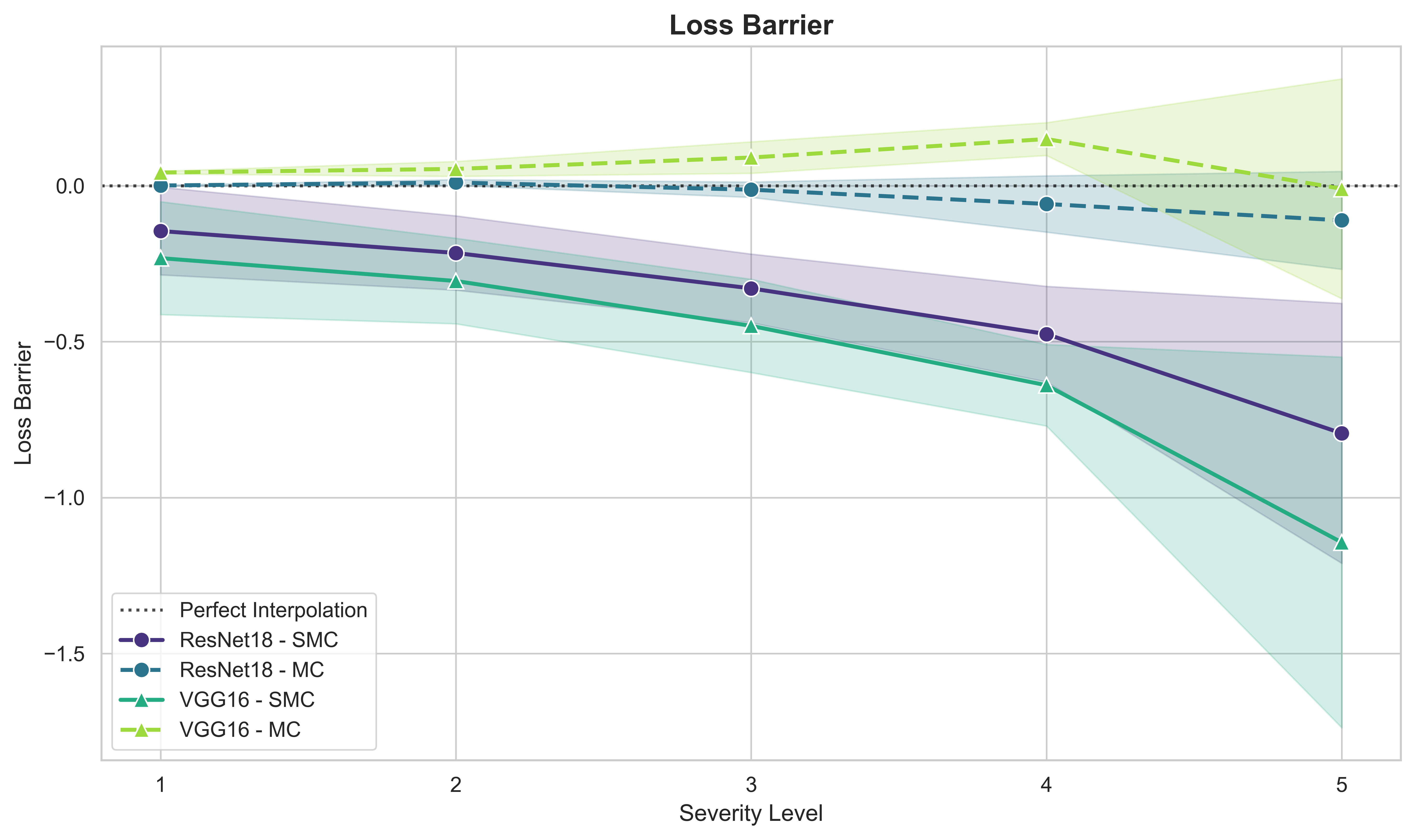}
	\caption{Loss barrier (averaged over blur types) vs.\ corruption severity for ResNet-18 and VGG16-BN on CIFAR-10-C. The SMC maintains deeply negative barriers for both architectures, while standard MC barriers remain near zero. VGG16-BN SMC barriers are consistently deeper than ResNet-18, suggesting purely sequential architectures benefit more from flat-basin connectivity. Shaded regions indicate $\pm 1$ std across blur types.}
	\label{fig:loss_barrier}
\end{figure}

Fig.~\ref{fig:loss_barrier} plots the loss barrier~\eqref{eq:loss_barrier} averaged across blur types as severity increases for both ResNet-18 and VGG16-BN. For both architectures, the standard MC barrier remains near zero or trends slightly positive, indicating that the narrow connecting valley collapses under distribution shift. Remarkably, the SMC barrier plunges into deeply negative territory: ResNet-18 reaches $-0.47$ and VGG16-BN reaches $-0.90$ at Severity~5, demonstrating that the optimized midpoint achieves lower loss than the average of the independently trained endpoint losses under corruption. We note that because the barrier is defined relative to the endpoint average (Eq.~\ref{eq:loss_barrier}), a negative value does not guarantee that the midpoint outperforms both endpoints individually; however, the consistently large negative magnitudes across corruption types suggest that the midpoint occupies a genuinely more robust region of the loss landscape. The deeper VGG16-BN barriers suggest that purely sequential architectures benefit even more from flat-basin connectivity, plausibly because the absence of residual skip connections produces sharper ridges between independently trained solutions~\cite{li2018visualizing}, making sharpness-aware regularization more impactful. The wider confidence bands at high severity indicate greater variability across blur types.

Table~\ref{tab:sev5_barrier} reports per-corruption Severity 5 barriers for both architectures. Under Gaussian blur, the ResNet-18 SMC achieves $-1.3582$ vs.\ $+0.3793$ for standard MC, a profound geometric advantage. The VGG16-BN SMC replicates this pattern with deeply negative barriers across all corruption types, confirming architectural transferability.

\begin{table}[htbp]
	\centering
	\caption{Midpoint Loss Barrier at Severity 5: ResNet-18 and VGG16-BN}\label{tab:sev5_barrier}
	\renewcommand{\arraystretch}{1}
	\begin{tabular}{lcccc}
		\toprule
		& \multicolumn{2}{c}{\textbf{ResNet-18}} & \multicolumn{2}{c}{\textbf{VGG16-BN}} \\
		\cmidrule(lr){2-3} \cmidrule(lr){4-5}
		\textbf{Corruption} & \textbf{SMC} & \textbf{MC} & \textbf{SMC} & \textbf{MC} \\
		\midrule
		Gaussian Blur & $-1.3582$ & $+0.3793$ & $-1.7862$ & $-0.6252$ \\
		Defocus Blur & $-0.9065$ & $+0.2218$ & $-1.1946$ & $-0.3307$ \\
		Glass Blur & $-0.3064$ & $+0.5403$ & $-0.8349$ & $-0.1357$ \\
		Motion Blur & $-0.4056$ & $+0.3333$ & $-0.6576$ & $-0.0287$ \\
		\bottomrule
	\end{tabular}
\end{table}

\subsubsection{Interpolation Advantage Dynamics}

\begin{figure}[htbp]
	\centering
	\includegraphics[width=0.5\textwidth]{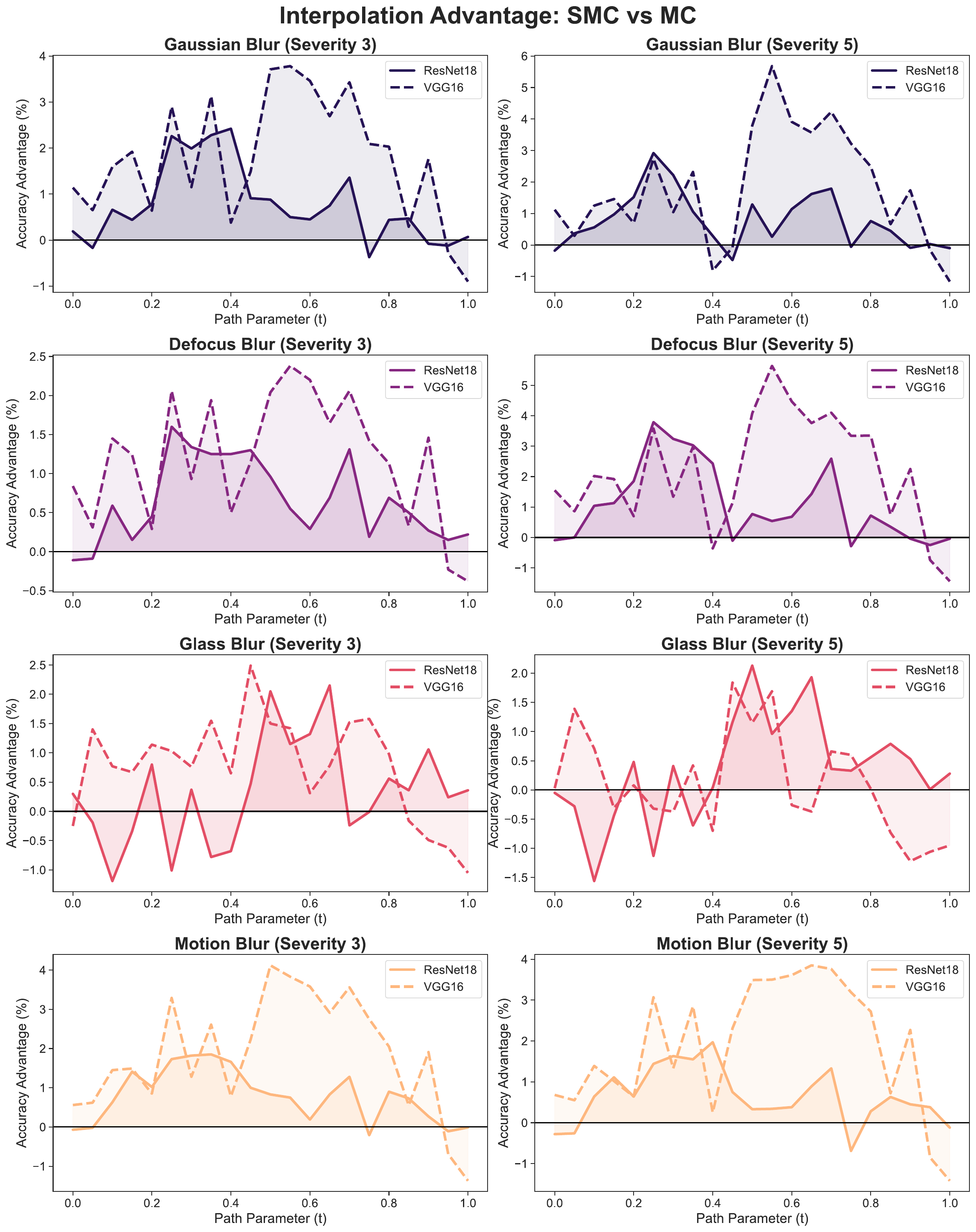}
    	\caption{Interpolation advantage ($\text{Acc}_\text{SMC} - \text{Acc}_\text{MC}$) along the full path for ResNet-18 (solid) and VGG16-BN (dashed) under Severity~3 and~5 across four blur types on CIFAR-10-C. Volatile positive spikes and quartiles reveal where standard MC is most fragile. VGG16-BN exhibits amplified advantages (up to $+6\%$ under Gaussian and Defocus blur), while Glass blur yields the smallest, noisiest gains for both architectures.}
	\label{fig:interp_adv}
\end{figure}

Fig.~\ref{fig:interp_adv} plots the pointwise accuracy advantage $\text{Adv}(t) = \text{Acc}_\text{SMC}(t) - \text{Acc}_\text{MC}(t)$ across the full path for Severity~3 and~5 under all four blur types. Volatile positive spikes at the midpoint and quartiles indicate regions where standard MC is most vulnerable under interpolation. VGG16-BN exhibits amplified advantages, peaking at $+6\%$ under Gaussian and Defocus blur, roughly double the ResNet-18 peaks. Glass blur yields the smallest, noisiest gains for both architectures.

\subsection{Ablation: Impact of SAM Perturbation Radius $\rho$}\label{subsec:rho_sweep}

\begin{figure}[htbp]
	\centering
	\includegraphics[width=0.85\linewidth]{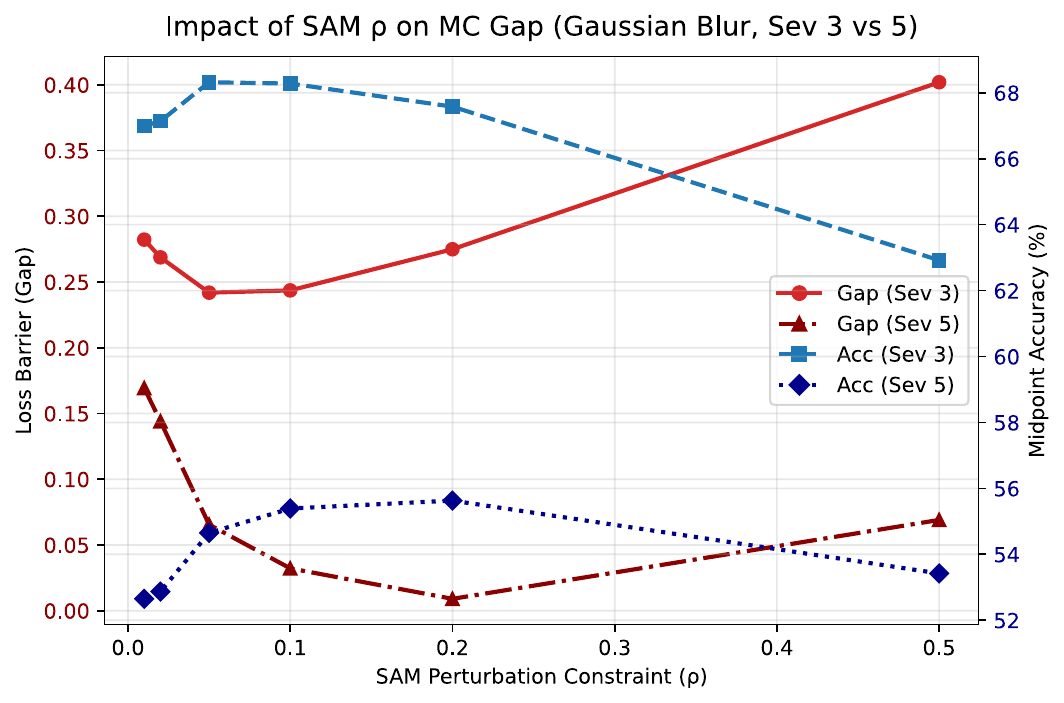}
	\caption{ResNet-18 dual-axis plot showing the loss barrier and midpoint accuracy across varying SAM perturbation radii $\rho$ under Gaussian blur (CIFAR-10-C). Increasing $\rho$ improves connectivity up to an optimal range, beyond which excessive noise collapses the path.}
	\label{fig:rho_sweep}
\end{figure}

The preceding OOD results demonstrate that the SMC framework consistently improves robustness, but the magnitude of this improvement depends on the SAM perturbation radius $\rho$. To isolate this effect, we sweep $\rho \in \{0.01, 0.02, 0.05, 0.1, 0.2, 0.5\}$ with frozen endpoints and evaluate under Gaussian blur at Severity 3 and 5 for both ResNet-18 and VGG16-BN. Fig.~\ref{fig:rho_sweep} summarizes the resulting trade-off between the loss barrier and midpoint accuracy for ResNet-18.

\begin{figure}[htbp]
    \centering
    \includegraphics[width=1\linewidth]{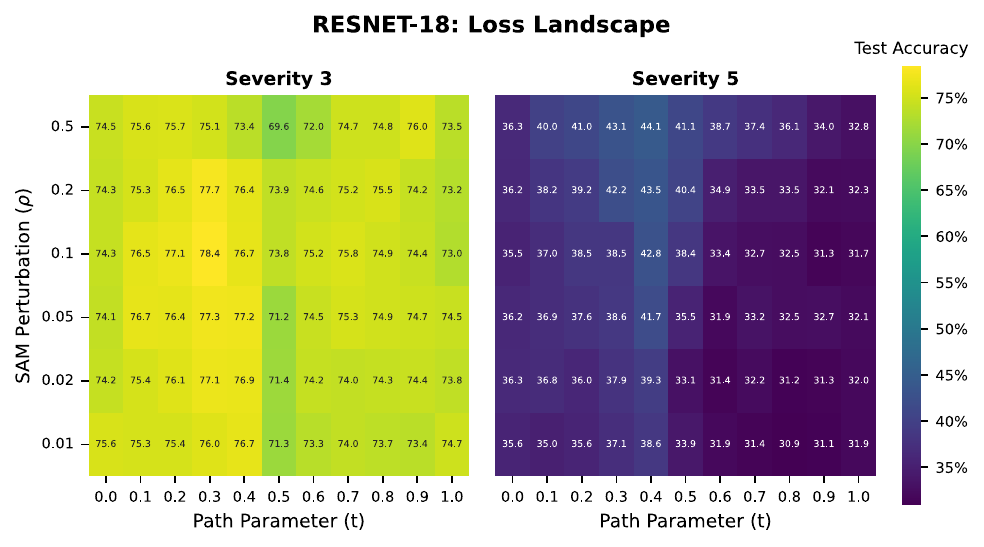}
    \caption{ResNet-18 accuracy landscape heatmaps across the path parameter ($t$) and SAM perturbation radius ($\rho$) under Gaussian blur. An optimal $\rho$ range is observed; excessive budgets collapse the path. The wider optimal range at Severity 5 confirms that stronger regularization is needed under more severe distribution shift.}
    \label{fig:rho_heatmaps_resnet}
\end{figure}

\begin{figure}[htbp]
    \centering
    \includegraphics[width=1\linewidth]{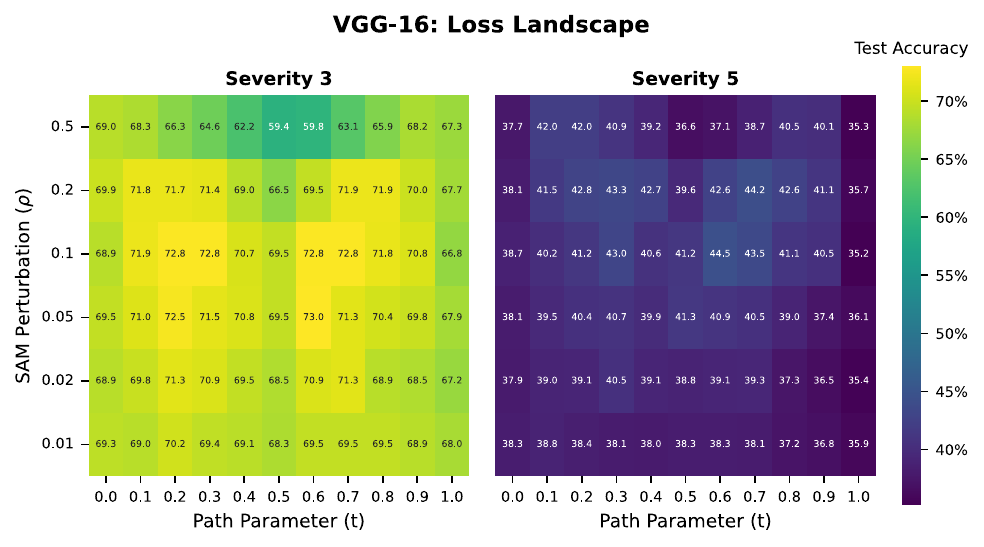}
    \caption{VGG16-BN accuracy landscape heatmaps across the path parameter ($t$) and SAM perturbation radius ($\rho$) under Gaussian blur. The pattern mirrors ResNet-18, confirming that the benefit is architecture-agnostic.}
    \label{fig:rho_heatmaps_vgg16}
\end{figure}

At Severity 3, increasing $\rho$ strictly improves the connectivity bridge up to $\rho=0.1$, after which the excessive noise injection collapses the path. At Severity 5, the optimizer requires stronger regularization, and increasing $\rho$ over a broader range consistently enhances performance. This pattern holds across both architectures (Figs.~\ref{fig:rho_heatmaps_resnet} and \ref{fig:rho_heatmaps_vgg16}), with moderate perturbation budgets ($\rho \approx 0.05$ to $0.1$) generating the optimal accuracy landscapes. These results confirm that the wide-valley constraint is critical under severe distribution shift and that the benefit is architecture-agnostic.

\subsection{Vision Transformer Connectivity}\label{subsec:vit}

Having established that SMC consistently benefits CNN architectures across corruption types and perturbation budgets, we now examine the computational cost and architectural implications for attention-based architectures. The clean connectivity results for ViT-Tiny on CIFAR-10 and ImageNet-100 were presented in Section~\ref{subsec:clean_conn} (Table~\ref{tab:clean_baseline} and Fig.~\ref{fig:vit_clean}), where the SMC consistently outperformed standard MC despite a residual midpoint barrier absent in CNNs.

Table~\ref{tab:vit_training_time} reports the computational cost. The SMC incurs a $5.6\times$ per-epoch overhead due to SAM's two-backpropagation structure, with 200 training epochs (vs.\ 50 for CNNs) to accommodate attention-based architectures.

\begin{table}[htbp]
	\centering
	\caption{ViT-Tiny Training Time and Computational Cost (CIFAR-10)}\label{tab:vit_training_time}
	\renewcommand{\arraystretch}{1.1}
	\resizebox{0.70\linewidth}{!}{
    \begin{tabular}{lcc}
		\toprule
		\textbf{Metric} & \textbf{SMC} & \textbf{MC} \\
		\midrule
		Time / epoch (s) & 39.41 & 7.03 \\
		Throughput (img/s) & 1{,}268.6 & 7{,}114.7 \\
		Training epochs & \multicolumn{2}{c}{200} \\
		Overhead factor & \multicolumn{2}{c}{$5.6\times$} \\
		\midrule
		\multicolumn{3}{l}{\textit{Hardware}: $4\times$ NVIDIA GH200 120GB, DDP} \\
		\multicolumn{3}{l}{\textit{CPU}: 288 logical cores} \\
		\bottomrule
	\end{tabular}
    }
\end{table}

One plausible contributing factor to the residual midpoint barrier in ViT-Tiny is the permutation symmetry of multi-head self-attention: query-key-value projections can be arbitrarily reordered across heads without changing the network function, so two independently trained ViTs may encode functionally equivalent but weight-space permuted representations. Cubic interpolation between such permuted representations may partially disrupt learned token-mixing structure. However, we have not empirically verified this conjecture; testing it by applying permutation-alignment techniques (e.g., Git Re-Basin~\cite{ainsworth2023git}, optimal transport) prior to path optimization is a promising direction for eliminating this residual barrier. A systematic OOD evaluation of ViT-Tiny under CIFAR-10-C corruptions, analogous to the CNN analysis in Section~\ref{subsec:ood}, remains an important direction for future work.

\FloatBarrier
\subsection{Discussion}\label{subsec:discussion}

The deeply negative loss barriers under distribution shift (Table~\ref{tab:sev5_barrier}) arise from asymmetric optimization: endpoints are trained with standard SGD without flatness constraints, while every interior point is optimized by $J_{\mathrm{sharp}}(\theta)$, which penalizes worst-case loss within a $\rho$-ball. Under distribution shift, the non-regularized endpoints degrade faster than the SAM-regularized interior, producing negative barriers that suggest the midpoint occupies a more robust region of the loss landscape than the endpoint average. An important caveat is that our framework guarantees \emph{weight-space} flatness ($\max_{\|\epsilon\|_2 \leq \rho} \loss(w + \epsilon) \approx \loss(w)$), which need not imply \emph{input-space} robustness ($\max_{\|\delta\|_\infty \leq \varepsilon} \loss(x + \delta; w)$). Practitioners should consider this distinction when deploying interpolated models in safety-critical applications.

\section{Conclusion}\label{sec:conclusion}
This paper proposed SMC, which integrates SAM's minimax objective into Mode Connectivity's path-finding formulation to discover parametric curves in weight space that are simultaneously low-loss and uniformly flat. Experiments across CNN (ResNet-18, VGG16-BN) and Transformer (ViT-Tiny) architectures on CIFAR-10 and ImageNet-100 show that SMC consistently produces flatter connectivity paths, yields meaningful accuracy gains under severe distribution shift, and discovers path-interior solutions that can outperform the average endpoint loss, as evidenced by deeply negative loss barriers. These results establish path-wise flatness as a practical geometric principle for robust generalization.

The current evaluation is limited to a single endpoint pair per architecture and does not ablate SAM-trained endpoints with standard MC or permutation-aligned linear interpolation. Future work includes adaptive perturbation scheduling along the curve, integration of permutation-alignment methods to reduce the residual ViT midpoint barrier, and evaluation of adversarial robustness along the optimized path using gradient-based attacks (e.g., PGD, FGSM~\cite{kim2020torchattacks}) to determine whether weight-space flatness confers any resistance to input-space perturbations.

\bibliographystyle{IEEEtran}
\bibliography{bibliography}

\end{document}